\documentclass[letterpaper, 10 pt, conference]{ieeeconf}  

\IEEEoverridecommandlockouts                              

\makeatletter
\let\NAT@parse\undefined
\makeatother

\usepackage[utf8]{inputenc}

\usepackage{tikz}
\usepackage{amsmath} 
\usepackage{amssymb}  
\usepackage{graphicx} 
\usepackage{algorithm}
\usepackage{subfigure}
\usepackage[export]{adjustbox}
\usepackage[noend]{algpseudocode}
\usepackage{placeins}
\usepackage{pifont}
\usepackage{moreverb}
\usepackage{caption}

\usepackage{url}
\usepackage[hidelinks]{hyperref}

\newcommand{\cmark}{\ding{51}}%
\newcommand{\xmark}{\ding{55}}%

\newcommand\copyrighttext{%
  \footnotesize \textcopyright 2019 IEEE. Personal use of this material is permitted.
  Permission from IEEE must be obtained for all other uses, in any current or future 
  media, including reprinting/republishing this material for advertising or promotional 
  purposes, creating new collective works, for resale or redistribution to servers or 
  lists, or reuse of any copyrighted component of this work in other works. 
  }
\newcommand\copyrightnotice{%
\begin{tikzpicture}[remember picture,overlay]
\node[anchor=south,yshift=10pt] at (current page.south) {\fbox{\parbox{\dimexpr\textwidth-\fboxsep-\fboxrule\relax}{\copyrighttext}}};
\end{tikzpicture}%
}

\title{\LARGE \bf
  Graduated Fidelity Lattices for Motion Planning under Uncertainty
}

\author{Adri\'an Gonz\'alez-Sieira$^{1}$, Manuel Mucientes$^{1}$ and Alberto Bugar\'in$^{1}$
\thanks{This research was supported by the Spanish Ministry of Economy and Competitiveness (grants TIN2014-56633-C3-1-R and TIN2017-84796-C2-1-R), and the Galician Ministry of Education, Culture and Universities (grants GRC2014/030 and accreditation 2016-2019, ED431G/08). These grants are co-funded by the European Regional Development Fund (ERDF/FEDER program).}
\thanks{$^{1}$Authors are with Centro Singular de Investigación en Tecnoloxías da Información (CiTIUS), Universidade de Santiago de Compostela, Spain.
        {\tt\small \{adrian.gonzalez, manuel.mucientes, alberto.bugarin.diz\}@usc.es}}%
}

\algnewcommand{\IfThenElse}[3]{
  \State \algorithmicif\ #1\ \algorithmicthen\ #2\ \algorithmicelse\ #3}

\begin{document}
\maketitle
\copyrightnotice
\thispagestyle{empty}
\pagestyle{empty}

\begin{abstract}

In this work we present a state lattice based approach for motion planning in mobile robotics. Sensing and motion uncertainty are managed at planning time to obtain safe and optimal paths. To do this reliably, our approach estimates the probability of collision taking into account the robot shape and the uncertainty in heading. We also introduce a novel graduated fidelity approach and a multi-resolution heuristic which adapt to the obstacles in the map, improving the planning efficiency while maintaining its performance. Results for different environments, shapes and motion models are reported, including experiments with real robots.

\end{abstract}

\section{Introduction}
\label{sec:introduction}
Motion planning in mobile robotics has been successfully addressed using stochastic and deterministic sampling strategies \cite{LaValle2006}. Among the latter, state lattices are noteworthy for their regularity and for having the structure of a graph in which the discrete states are connected by motions extracted from the dynamics model. Thus, optimal paths satisfying the kinematic constraints can be found with a search algorithm.

Managing motion and sensing uncertainty is also important, because the safety of the planned paths is crucial in real world applications. Doing so at planning time allows selecting the best path according to its probability of collision, which should be reliably estimated taking into account the robot dimensions and the uncertainty in heading.

The fidelity of the lattice is the resolution of the sampled states. Decreasing the fidelity improves the planning efficiency, but it affects the optimality and the capacity to obtain valid solutions. However, in a graduated fidelity lattice the resolution can vary and adapt to the obstacles in the map, managing the trade off between planning performance and efficiency. Heuristics are key for the latter, so reducing their computation time is equally relevant. Multi-resolution maps, like octrees \cite{Hornung2013_AR}, also adapt  to the obstacles, so they can be a good source of information for the fidelity selection and to introduce multi-resolution techniques in the heuristics.

In this paper we present a state lattice based motion planner that manages the uncertainty at planning time. Our main contributions are: 1) a sampling strategy which reliably estimates the probability of collision of the robot taking into account its dimensions and the uncertainty in heading; 2) a novel graduated fidelity approach which, unlike prior works, adapts to the maneuverability of the robot and to the obstacles in the map; and 3) a novel multi-resolution heuristic with improved efficiency and scalability in large environments. This approach allows to reduce significantly the runtime of the planner while maintaining its performance.

\section{Related work}
Sampling based techniques combined with search algorithms have been successfully applied to the field of motion planning. Random sampling ---Probabilistic Roadmaps (PRM) \cite{Kavraki1996}, Rapidly Exploring Random Trees (RRT) \cite{LaValle2001}--- and deterministic sampling techniques have been described in the literature. Among the latter, state lattices \cite{Pivtoraiko2005} benefit from their regularity to pre-compute a set of canonical actions from the dynamics model. Graduated fidelity state lattices were introduced by \cite{Pivtoraiko2008_IROS}, although this approach does not take into account the dynamics in the low fidelity areas, thus making impossible managing the uncertainty at planning time. In \cite{Likhachev2009} this issue was addressed using a subset of the motion primitives to connect the states within low fidelity areas. However, the approach of \cite{Likhachev2009} and \cite{Pivtoraiko2009} is to obtain a pre-planned path and improve it in real time as the robot moves along the planned path, increasing the fidelity of the lattice around its position. Combining this with uncertainty management is not trivial because every time the lattice changes the uncertainty of the affected paths has to be re-computed, which is a costly operation. Instead, our graduated fidelity approach only depends on the obstacles in the map and the maneuverability of the robot, addressing these issues.

Different heuristics to improve the planning efficiency were described in the literature. In \cite{Knepper2006} they presented an admissible heuristic ---FSH--- which copes with the robot dynamics. Although its obtention is computationally expensive, it can be pre-computed offline and stored in a look-up table. In \cite{Likhachev2009} they introduced H2D, a low-dimensional heuristic which takes into account the obstacles in the map, and which combined with FSH obtained good results. However, it is computed applying the Dijkstra's algorithm over a grid with a fixed resolution, so its efficiency when dealing with large and uncluttered environments can be improved. For this reason we introduce H2DMR, a heuristic based on that of \cite{Likhachev2009}, which relies on a multi-resolution grid and improves the planning efficiency and scalability.

Regarding the uncertainty management, prior works used the theory of Markov Decision Processes ---MDPs and POMDPs--- to deal with the inaccuracies in the controls \cite{Alterovitz2007} and measurements \cite{Kaelbling1998_AI}, although they had scalability issues \cite{Papadimitriou1987}. These were addressed by \cite{VanDenBerg2012_IJRR}, which computes a locally-optimal solution given an initial path. However, this approach can only be applied to smooth dynamics and observations. Other works, like \cite{VanDenBerg2011} and \cite{Bry2011}, use the Extended Kalman Filter ---EKF--- to deal with motion and sensing uncertainty without assuming maximum likelihood observations. \cite{VanDenBerg2011} relies on RRT to compute a set of paths and selects among them the one with the minimal probability of collision, although this does not guarantee finding a good solution in all cases. \cite{Bry2011} addresses this drawback growing a search tree which predicts the uncertainty for all candidate paths. Despite their good results in uncertainty management, both approaches estimate the probability of collision using simplified versions of the robot shape. \cite{VanDenBerg2011} approximates the robot by its bounding circle, while \cite{Bry2011} checks collisions with the predicted distributions. This affects the reliability of the planner and leads to potential failures when the shape is far from these approximations ---e.g. asymmetric or irregular shapes. Our approach considers the real robot shape with a deterministic sampling strategy which computes the probability of collision from the predicted uncertainty, also taking into account the uncertainty in heading.

\section{Planning on state lattices}
\label{sec:problemformulation}
Our motion planner relies on a state lattice, so the state space, $\mathcal{X}$, is sampled in a regular manner ---a rectangular arrangement was used in this work, although others are possible. A canonical set of actions extracted from the dynamics model ---$\mathcal{U}$, also called motion primitives--- connects the lattice states, $\mathcal{X}_{lat}$. Due to their regular arrangement, these actions are position-independent and can be computed offline and replicated to generate the whole connectivity. Moreover, these actions respect the kinematic restrictions.

The motion primitives are obtained using numerical optimization, as detailed in \cite{Howard2007_IJRR}. The cost of the resulting actions is optimal given the constraints: the initial an final states, belonging to $\mathcal{X}_{lat}$, and the dynamics model.

Due to the graph structure of the state lattice, an informed search algorithm can be used to find the optimal path in it. These algorithms rely on heuristics, which estimate the cost between each state and the goal, to efficiently explore the state space. The planner we present uses Anytime Dynamic A* \cite{Likhachev2005}, AD*, due to its capability to obtain sub-optimal solutions and refine them iteratively without planning from scratch. The value of the heuristic is inflated by a parameter $\epsilon$, which acts as boundary for the cost of the solution.

As heuristic, we combine the proposed H2DMR ---which deals with the obstacles in the map, like H2D \cite{Likhachev2009}--- with FSH \cite{Knepper2006} ---which copes with the kinematic restrictions while considering free space---, as follows:
\begin{equation}
\text{HEURISTIC}(x) = \max \left( \textit{h2dmr}(x), \textit{fsh}(x) \right)
\end{equation}
This allows estimating the cost to the goal in an accurate manner, which benefits the planning efficiency.

\begin{algorithm}[b]

  \begin{algorithmic}[1]
    \Require $x^0$, initial state; $x^G$, goal state; initially $\epsilon = \epsilon_0$
    \Function{ main }{ $x^0$, $x^{G}, \epsilon$ }
        \State \label{alg:searchLoop:h2dmr} \Call {initializeHeuristic}{$x^0$, $x^G$} \Comment {Alg. \ref{alg:h2dmr}}
        \While {$\epsilon >= 1$}
            \State $c_{x^{0}} = 0$; $h_{x^{0}} = $ \Call{heuristic}{$x^{0}$}; $\textit{OPEN} = \{ x^{0} \}$
            \Repeat
                \State \label{alg:searchLoop:select} $x^a = \arg \min_{x \in \textit{\tiny OPEN}} (c_x + \epsilon \cdot h_x)$ 
                \State $X^b = $ \Call{successors}{$x^{a}$} \Comment{Alg.\ref{alg:multiresolution}} 
                \label{alg:searchLoop:multiresolution}
                \ForAll{ $x^b \in X^b$} 
                    \State $c_{x^{b}} = c_{x^{a}} + $ \Call{cost}{$x^a, x^b$}  \Comment{Alg. \ref{alg:probabilityCollision}} \label{alg:searchLoop:evaluate}
                    \State $h_{x^{b}} = $ \Call{heuristic}{$x^{b}$}  \Comment{Alg. \ref{alg:h2dmr}} \label{alg:searchLoop:heuristic}
                    \State $\textit{OPEN} = \textit{OPEN} \cup \{x^{b} \}$
                    \label{alg:searchLoop:insertOpen}
                \EndFor
            \Until{ $x^a = x^{G}$ }
            \State publish $\textit{path}(x^0, x^a)$ and decrease $\epsilon$
            \label{alg:searchLoop:decreaseEpsilon}
        \EndWhile
        \State \Return
    \EndFunction
  \end{algorithmic}
 
  \caption{Main operations of the search algorithm}
  \label{alg:searchLoop}
  
\end{algorithm}

Alg. \ref{alg:searchLoop} summarizes the operations done by AD*. Inputs ---$x^0$ and $x^G$--- are the initial state and the goal, and the output is the path with the minimal cost connecting them. H2DMR is initialized before running the planner ---Alg. \ref{alg:searchLoop}:\ref{alg:searchLoop:h2dmr}---, since it depends on the location of the goal and the obstacles in the environment. The algorithm is then run iteratively, starting with $\epsilon = \epsilon_0$ and decreasing the value of the parameter until the optimal solution is found ---Alg. \ref{alg:searchLoop}:\ref{alg:searchLoop:decreaseEpsilon}.

Given a value for $\epsilon$, the algorithm iteratively extracts a state $x^a$ from the $\textit{OPEN}$ queue ---Alg. \ref{alg:searchLoop}:\ref{alg:searchLoop:select}. This is the state minimizing the added cost from the start and the estimated one to the goal ---$c_x$ and $h_{x}$, respectively. The latter is given by the heuristic and scaled by $\epsilon$. The successors for $x^a$, $X^b$, are then obtained in Alg. \ref{alg:searchLoop}:\ref{alg:searchLoop:multiresolution} and the cost of each transition is evaluated ---Alg. \ref{alg:searchLoop}:\ref{alg:searchLoop:evaluate}. Finally, each state $x^b$ is inserted in $\textit{OPEN}$ after computing its heuristic ---Alg. \ref{alg:searchLoop}:\ref{alg:searchLoop:heuristic}-\ref{alg:searchLoop:insertOpen}. A valid path is found when $x^a$ is the goal. 

To ensure the safety of the planned paths they have to be evaluated in terms of probability of collision. This requires managing the uncertainty at planning time, for which we use the approach of \cite{Bry2011} to predict the Probability Density Functions of the robot being in each state of the path ---PDFs. This method focuses on nonlinear, partially observable systems with dynamics and observations described in a discrete time manner:
\begin{equation}
    \begin{aligned}
    & x_{t+1} = f(x_{t}, u_{t}) + m_t, &\qquad m_t \sim \mathcal{N}(0, M_t)
    \\
    & z_{t} = z(x_{t}) + n_t , &\qquad n_t \sim \mathcal{N}(0, N_t)
    \end{aligned}
\end{equation}
where $x_t \in \mathcal{X}$, $u_t \in \mathcal{U}$ and $z_t$ are the robot states, the controls and the measurements. $m_t$ and $n_t$ are Gaussian motion and observation noises with covariances $M_t$ and $N_t$. The PDFs are estimated considering the influence of a Linear Quadratic Gaussian \cite{Bertsekas1995} controller ---LQG--- when executing the planned paths, using linearized versions of $f$ and $z$. The uncertainty depends on that at the initial state, the controls and the accuracy of the measurements; and it is therefore different for each candidate path in the lattice. 

\section{Reliable probability of collision}
\label{sec:prob-collision}
The distributions resulting from predicting the uncertainty are used to estimate the probability that the robot collides when executing the paths. We introduce a novel method to do so in a reliable manner, taking into account the real shape of the robot and dealing with the uncertainty in heading.

The goal of the planner is to obtain paths minimizing the probability of collision and the traversal time. This is achieved with a cost function which evaluates a path between $x^a$ and $x^b$ with three objectives: a safety measure, which depends on the probability of collision; the traversal time; and the uncertainty at $x^b$ ---$c^{a:b}$,  $t^{a:b}$ and $\Sigma^{b}$, respectively. An order of priority is introduced in these elements, so that the planner first minimizes the probability of collision, then the traversal time and finally the uncertainty at the goal.

\begin{algorithm}[b!]

  \caption{Cost of a trajectory between $x^a$ and $x^b$}
  \label{alg:probabilityCollision}

  \begin{algorithmic}[1]
    \Require $x^a$ and $x^b$, beginning and final states
    \Function{cost}{$x^a, x^b$}
        \State $P^{a:b} = $ \Call{uncertainty}{$x^a$, $x^b$} \Comment{PDF prediction}
         \label{alg:probabilityCollision:uncertainty}
        \State $t^{a:b} = \textit{time}(u^{a:b})$ \label{alg:probabilityCollision:initialize-2}
        \State $c^{a:b} = 0$
        \ForAll { $x_t^{a:b} \sim \mathcal{N}(\bar{x}_t^{a:b}, \Sigma_t^{a:b}) \in P^{a:b}$ }
        	\State $w_c = 0$; $w_t = 0$ \label{alg:probabilityCollision:collision-begin}
        	\State $X^S = \textit{sampling}(x_t^{a:b})$ \label{alg:probabilityCollision:sampling}
        	\ForAll { $x^s \in X^S$ } \label{alg:probabilityCollision:for}
        	    \State $w_t = w_t + \textit{pdf}(x^s, x_t^{a:b})$ \label{alg:probabilityCollision:weight-total}
        	    \If {$\textit{collision}(x^s)$} \Comment{with real shape} \label{alg:probabilityCollision:if-collides}
        	        \State $w_c = w_c + \textit{pdf}(x^s, x_t^{a:b})$ \label{alg:probabilityCollision:collision-end}
    	        \EndIf
        	\EndFor
        	\State $c^{a:b} = c^{a:b} - \log(1 - w_c / w_t)$ \label{alg:probabilityCollision:safetycost}
          \EndFor
        \State \Return $\left[ c^{a:b}, t^{a:b}, \textit{tr}(\Sigma^{b}) \right]$ \Comment{$\Sigma^{b}$, uncertainty at $x^b$} \label{alg:probabilityCollision:return}
    \EndFunction
  \end{algorithmic}
  
\end{algorithm}

Alg. \ref{alg:probabilityCollision} details the evaluation of a path between two states $x^a$ and $x^b$. First, the PDFs of the path are predicted with the approach of Bry et. al. \cite{Bry2011} ---Alg.  \ref{alg:probabilityCollision}:\ref{alg:probabilityCollision:uncertainty}---, since they are needed to obtain $c^{a:b}$ and $\Sigma^{b}$. Conversely, $t^{a:b}$ is directly given by the motion primitives ---Alg.  \ref{alg:probabilityCollision}:\ref{alg:probabilityCollision:initialize-2}.

\begin{figure}[!b]
    \centering
    \includegraphics[width=0.7\columnwidth]{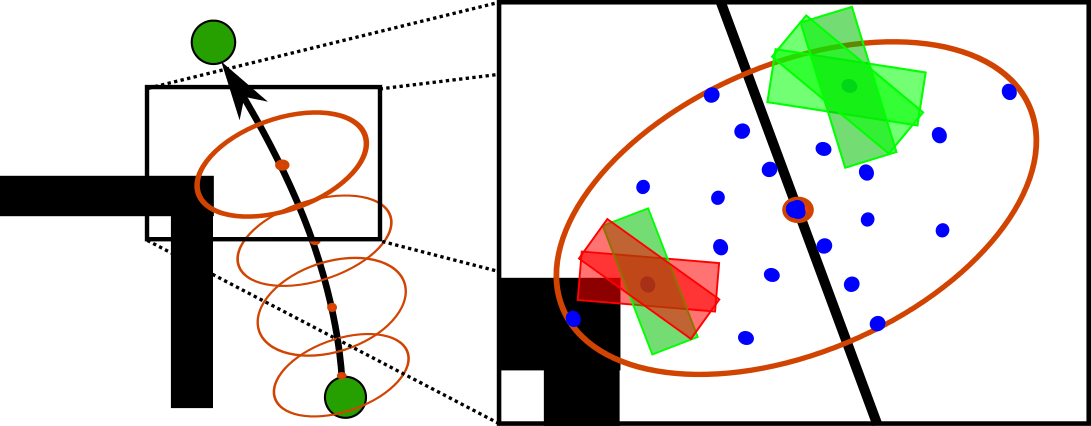}
    \caption{We estimate the probability of collision considering the robot real shape, deterministically sampling the PDFs.}
    \label{fig:ukf-sampling}
\end{figure}

The probability of collision is estimated from the PDFs with a deterministic sampling strategy, which is similar to the obtention of the sigma points in an Unscented Kalman Filter ---UKF \cite{Julier1995_ACC}. Given a $n$-dimensional distribution ---$x_t \sim \mathcal{N} ( \bar{x}_t, \Sigma_t)$--- a set of samples $X^S$ is obtained:
\begin{equation}
    \begin{aligned}
      & x_{i^+}^s, x_{i^-}^s = \bar{x}_t \pm \lambda \cdot \left( \sqrt{ \Sigma_t } \right)_i \qquad \qquad \ i=1, ..., n
  \end{aligned}
  \label{eq:sigmapoints}
\end{equation}
where $\left( \sqrt{ \Sigma_t } \right)_i$ is the $i$-th column of the factorized covariance matrix, and $\lambda$ allows obtaining samples with different distances to $\bar{x}_t$. Then, the following samples outside the main axes of the distribution are added to $X^S$:
\begin{equation}
    \begin{aligned}
      & x_{i^-j^+}^s, x_{i^-j^-}^s = x_{i^-}^s \pm \lambda \cdot \left( \sqrt{ \Sigma_t } \right)_j \qquad i, j=1, ..., n
      \\
      & x_{i^+j^+}^s, x_{i^+j^-}^s = x_{i^+}^s \pm \lambda \cdot \left( \sqrt{ \Sigma_t }
      \right)_j \qquad j \neq i
  \end{aligned}
\end{equation}
By doing so we achieve a better coverage of the PDF for collision check purposes ---Fig. \ref{fig:ukf-sampling}.

Each sample $x^s \in X^S$ is weighted according to its probability in the PDF. The probability that the robot collides ---Alg. \ref{alg:probabilityCollision}:\ref{alg:probabilityCollision:for}-\ref{alg:probabilityCollision:collision-end}--- is the ratio of the weights of the colliding samples ---checked with the real shape of the robot--- and the total weight. Then, the collision cost of the path, $c^{a:b}$, is obtained combining the estimations of the PDFs belonging to it, assuming they are independent ---Alg. \ref{alg:probabilityCollision}:\ref{alg:probabilityCollision:safetycost}. The cost due to $\Sigma^b$ is the trace of the matrix ---Alg. \ref{alg:probabilityCollision}:\ref{alg:probabilityCollision:return}.

The elements of the cost are compared hierarchically, which allows obtaining paths that are optimal in terms of safety, in the first place, then in terms of traversal time ---given the set of motion primitives used for planning---, and finally in terms of estimated final uncertainty.

\section{Efficient motion planning}
\label{subsection:multiresolution}
The fidelity of the lattice has a great impact in the planning efficiency. High fidelities allow representing more precisely the state space, while lower ones benefit the runtime at the expense the cost of the paths. The trade off between them can be properly managed with a graduated fidelity lattice, specially if it adapts to the obstacles in the map.

\begin{figure}[!b]
    
    \subfigure[]{\includegraphics[width=0.465\columnwidth, valign=t]{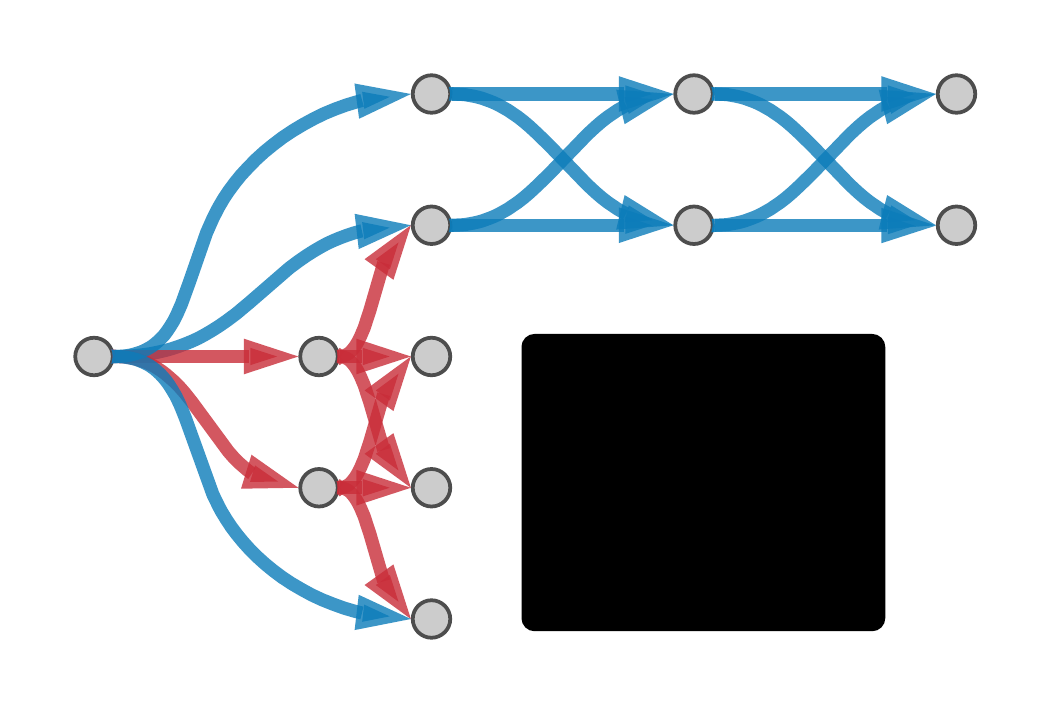}
    \vphantom{\includegraphics[width=0.465\columnwidth, valign=t]{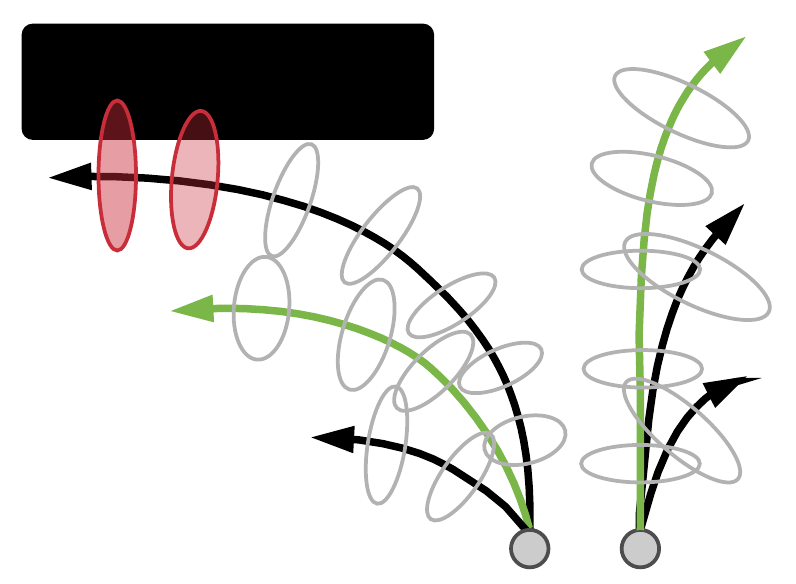}}%
    \label{fig:gf-lattice}}%
    \hfill%
    \subfigure[]{\includegraphics[width=0.465\columnwidth, valign=t]{graduated-fidelity} \label{fig:graduated-fidelity}}
    
    \caption{Graduated fidelity approach. (a) shows a lattice combining high (red) and low (blue) fidelity; (b) the fidelity selection approach: the longest collision free motion of each group is selected (green).}

\end{figure}

The standpoint of the proposed method is to use high fidelity only in those areas which require complex maneuvering ---i.e. near the obstacles---, selecting long trajectories whenever possible ---Fig. \ref{fig:gf-lattice}. Thus, the state space is simplified reducing the density of lattice states and candidate paths connecting them. This is achieved grouping together similar actions in $\mathcal{U}$ and selecting only one representative of each group when generating the successors of a state ---Alg. \ref{alg:searchLoop}:\ref{alg:searchLoop:multiresolution}. The selected maneuver is the longest one which does not affect the probability of collision ---Fig. \ref{fig:graduated-fidelity}. Each group contains those trajectories beginning and ending with the same heading and velocities, $\mathcal{U}_{(\theta_i, v_i, \omega_i, \theta_f, v_f, \omega_f)}$, so that it has maneuvers of the same kind but different length. Any two groups are disjoint, while the union of all of them is the whole set of motion primitives, $\mathcal{U}$.

\begin{algorithm}[t!]

  \begin{algorithmic}[1]
    \Require $\mathcal{U} = \{ \mathcal{U}_{(\theta_i, v_i, \omega_i, \theta_f, v_f, \omega_f)} \}, \forall (\theta, v, \omega) \in \mathcal{X}_{lat}$
    \Function{successors}{$x^a$}
        \State $\theta_i = x^a_\theta$; $v_i = x^a_v$; $\omega_i = x^a_\omega$; $\Gamma = \emptyset$
        \For{ $U \in \{ \mathcal{U}_{ ( \theta_i, v_i, \omega_i, \theta_f, v_f, \omega_f) } \}, \forall (\theta_f, v_f, \omega_f)$ } \label{alg:multiresolution:groups}
            \Repeat \label{alg:multiresolution:repeat-begin}
                \State $U^c = \arg \max_{t^{a:b}}(U)$ \Comment {Get longest} \label{alg:multiresolution:longest}
                \State $U = U \setminus U^c$
            \Until { \Call{check}{$U_{x^a}^c, U_{x^b}^c$} $\vee\ U == \emptyset$ } \label{alg:multiresolution:repeat-end}
            \State $\Gamma = \Gamma \cup U^c$ \label{alg:multiresolution:neighbor}
        \EndFor
        \State \Return $\Gamma$
    \EndFunction
    
    \Function{check}{$x^a$, $x^b$} \label{alg:multiresolution:function-check}
        \State $\kappa^{a} = \textit{cell}(x^{a})$; $\kappa^{b} = \textit{cell}(x^{b})$ \Comment {Get map cells} \label{alg:multiresolution:size-begin}
        \State $s^{a} = \textit{size}(\kappa^{a})$; $s^{b} = \textit{size}(\kappa^{b})$ \Comment {Get size of cells} \label{alg:multiresolution:size-end}
        \State $c^{a:b} = $ \Call{cost}{$x^{a}, x^{b}$}$[0]$ \Comment {Alg. \ref{alg:probabilityCollision}} \label{alg:multiresolution:cost}
        \State \Return { $( s^{a} + s^{b} \geqslant \left \| x^{a} - x^{b} \right \| ) \wedge ( c^{a:b} == 0 ) $ } \label{alg:multiresolution:function-check-return}
    \EndFunction
  \end{algorithmic}
  
  \caption{State successors in a graduated fidelity lattice}
  \label{alg:multiresolution}
  
\end{algorithm}

Alg. \ref{alg:multiresolution} details this process. The successors of a state $x^a$, $\Gamma$, are selected from those groups of trajectories with the same initial heading and velocities ---Alg. \ref{alg:multiresolution}:\ref{alg:multiresolution:groups}. For each group, all candidates are evaluated in descending order by $t^{a:b}$, and the first one fulfilling the restrictions is selected ---Alg. \ref{alg:multiresolution}:\ref{alg:multiresolution:repeat-begin}-\ref{alg:multiresolution:repeat-end}. These restrictions ---Alg. \ref{alg:multiresolution}:\ref{alg:multiresolution:function-check}-\ref{alg:multiresolution:function-check-return}--- relate to the resolution of the source and destination map cells ---$s^a$ and $s^b$--- and the probability of collision. The goal of the former is to choose maneuvers in accordance with the resolution of the map, which is a good indicator of the complexity of the environment. The latter maintains the safety of the solutions and it is obtained from the cost function ---Alg. \ref{alg:probabilityCollision}.

With this approach the density of states and maneuvers significantly decreases except near the obstacles, where complex maneuvering is required. This improves the planning efficiency, while the cost of the solutions is barely affected.

\subsection{Multi-resolution heuristic}
\label{subsection:heuristics}
Heuristics play a key role in the planning efficiency and their obtention time should be the lowest possible. Unlike FSH, H2D cannot be computed offline, since it depends on the obstacles in the map. Therefore, we propose H2DMR, a multi-resolution version of the latter which takes advantage of the octree map to reduce its obtention time.

Dijkstra's algorithm is used to build a multi-resolution grid which stores the distance between each point and the goal considering the obstacles ---Alg. \ref{alg:h2dmr}. Collisions are checked with the inscribed circle to the real shape to maintain the optimistic nature of the estimations. The resolution of this grid depends on that of the map and the highest fidelity of the motion primitives, $f^+$. H2DMR uses positions instead of states, and the cost is the euclidean distance ---Alg. \ref{alg:h2dmr}:\ref{alg:h2dmr:cost:begin}-\ref{alg:h2dmr:cost:end}. Starting in the goal, $p^G$, the point with the lowest cost from the start is iteratively selected ---Alg. \ref{alg:h2dmr}:\ref{alg:h2dmr:select}--- and its successors explored. As in H2D \cite{Likhachev2009}, the stopping condition is reaching twice the cost between $p^0$ and $p^G$.

\begin{algorithm}[t!]
  \begin{algorithmic}[1]
    \Require $f^+$, highest fidelity of the lattice
    \Function{initializeHeuristic}{$x^0, x^G$} \label{alg:h2dmr:calculate:begin}
        \State $p^0 = \textit{position}(x^0)$ ; $p^G = \textit{position}(x^G)$; 
        \State $c(p^G) = 0$; $\textit{OPEN} = \{ p^G \}$
        \Repeat
            \State $p = \arg \min_{p \in \textit{\tiny OPEN}} c(p)$ \label{alg:h2dmr:select}
            \State $\kappa = \textit{cell}(p)$ \Comment {Get map cell containing $p$} \label{alg:h2dmr:select-cell}
            \ForAll { $\kappa^{\prime} \in \textit{adjacent}(\kappa)$ } \label{alg:h2dmr:adjacent} \Comment{Adjacents to $\kappa^{\prime}$}
                \If {$\textit{size}( \kappa^{\prime} ) > f^+$} \label{alg:h2dmr:if-1-begin} \Comment {Size of cell $\kappa^\prime$}
                    \ForAll{ $\kappa^{\prime\prime} \in \textit{subcells}(\kappa^{\prime})$} 
                        \State $p^{\prime} = \textit{position}(\kappa^{\prime\prime})$ \label{alg:h2dmr:coord-1} \Comment{Center of $\kappa^{\prime\prime}$}
                        \State $c(p^{\prime}) = c(p) + $ \Call{costH}{$p, p^{\prime}$}
                        \State $OPEN = OPEN \cup \{ p^{\prime} \}$ \label{alg:h2dmr:if-1-end}
                    \EndFor
                \Else \label{alg:h2dmr:if-2-begin}
                    \State $\kappa^{\prime\prime} = \textit{adjust}(\kappa^{\prime}, f^+)$ \Comment{Adjust size to $f^+$}
                    \State $p^{\prime} = \textit{position}(\kappa^{\prime\prime})$ \label{alg:h2dmr:coord-2} \Comment{Center of $\kappa^{\prime\prime}$}
                    \State $c(p^{\prime}) = c(p) + $ \Call{costH}{$p, p^{\prime}$}
                    \State $OPEN = OPEN \cup \{ p^{\prime} \}$ \label{alg:h2dmr:if-2-end}
                \EndIf
            \EndFor
        \Until{ $c(p) > 2 \cdot c(p^0)$ } \label{alg:h2dmr:stop}
    \EndFunction \label{alg:h2dmr:calculate:end}
    
    \Function{ costH }{ $p$, $p^{\prime}$ } \Comment{Uses the optimistic shape} \label{alg:h2dmr:cost:begin}
        \IfThenElse{$\textit{collides}(p^{\prime})$}{\Return $\infty$;}{\Return $\|p^{\prime} - p\|$}  \label{alg:h2dmr:collision}
    \EndFunction \label{alg:h2dmr:cost:end}
  \end{algorithmic}
  
  \caption{Obtention of H2DMR}
  \label{alg:h2dmr}
  
\end{algorithm}

\begin{figure}[t!]
    \centering
    \includegraphics[width=0.9\columnwidth]{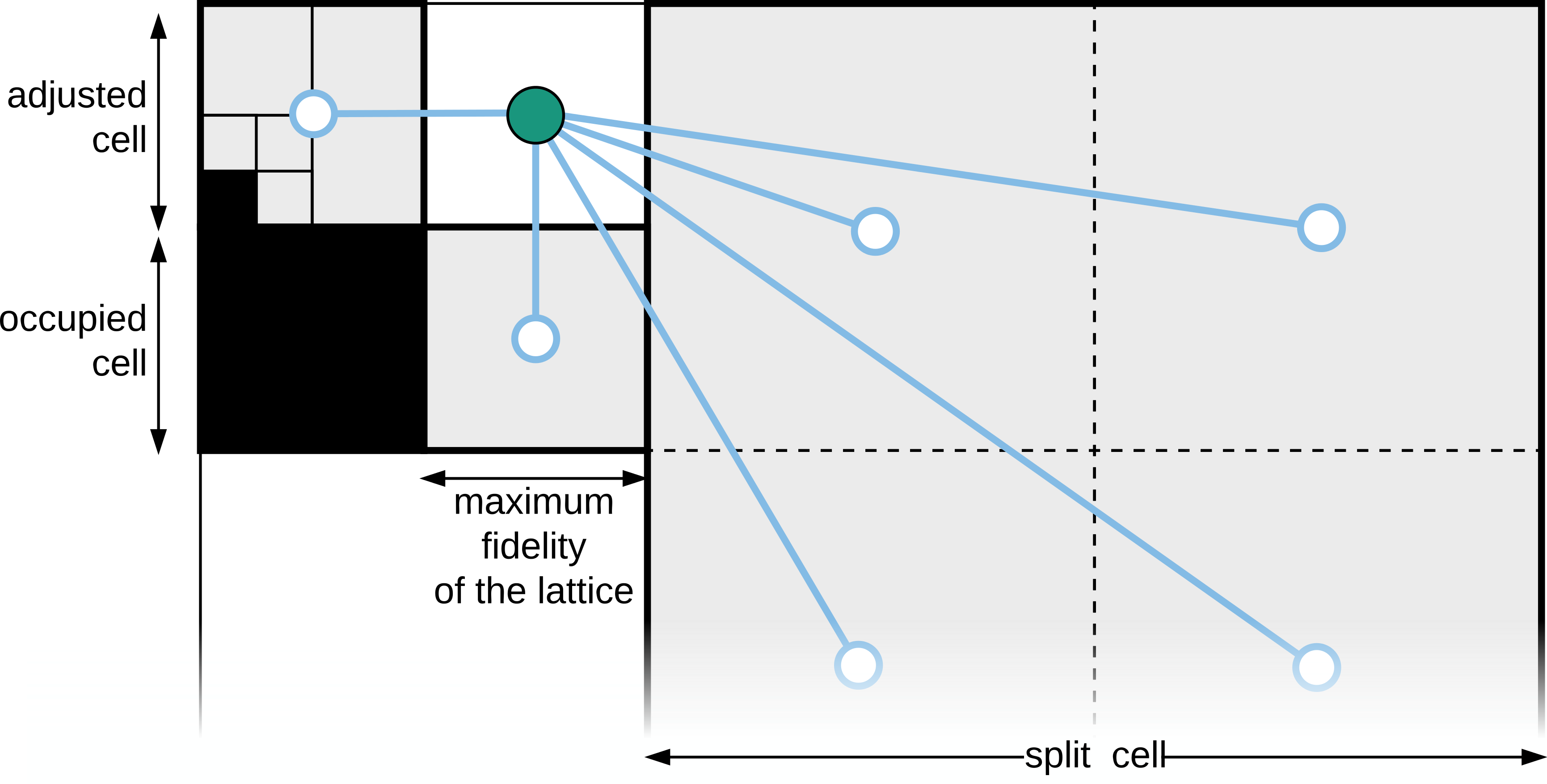}
    \caption{Given a point (in green), its neighbors (in blue) are the adjacent map cells (in gray). Those bigger than $f^+$ are split, otherwise bigger ones containing them are selected.}
    \label{fig:h2dmr}
\end{figure}

This approach takes advantage of the octree structure ---Fig. \ref{fig:h2dmr}. Given a point $p$, its successors are the center of the map cells which are adjacent to the one containing $p$. If the size of a cell is higher than $f^+$, it is split into subcells to keep the accuracy of the heuristic ---Alg. \ref{alg:h2dmr}:\ref{alg:h2dmr:if-1-begin}-\ref{alg:h2dmr:if-1-end}. Otherwise a bigger cell containing it is selected ---Alg. \ref{alg:h2dmr}:\ref{alg:h2dmr:if-2-begin}-\ref{alg:h2dmr:if-2-end}--- to limit the highest resolution of the grid to $f^+$.

AD* uses the costs stored in this multi-resolution grid as heuristic. However, the positions of H2DMR and the lattice states do not directly match, so the value for a state $x^a$ is given by the position, $p$, contained in the cell of $x^a$ or one of its adjacents which minimizes the sum of its cost in the grid ---$c(p)$--- and its distance to $x^a$:
\begin{equation}
    \textit{h2dmr}(x^a) = \arg \min_p (\lVert x^a - p \rVert + c(p) )
\end{equation}

This approach outperforms H2D in the obtention time of the grid, since it requires significantly less iterations to build it. Moreover, due to the octree structure the benefits are even more noticeable in large and uncluttered environments.

\section{Results}
\label{sec:results}
The following experiments were run on a computer with a CPU Intel Core™ i7-4790 and 16 GB of RAM. Motion uncertainty is $M_t=0.01 \cdot I$. Measurement noise is $N_t=0.01 \cdot I$ or $N_t=\infty \cdot I$, depending on whether the robot is in a location denied area or not.

\subsection{Reliable probability of collision}
In this section we report results for a robot with 2D Ackermann dynamics and a set of 336 motion primitives with lengths between $0.5\ m$ ---$f^+$--- and $8\ m$.

\begin{figure}[b]
    \subfigure[]{\includegraphics[height=4.8cm]{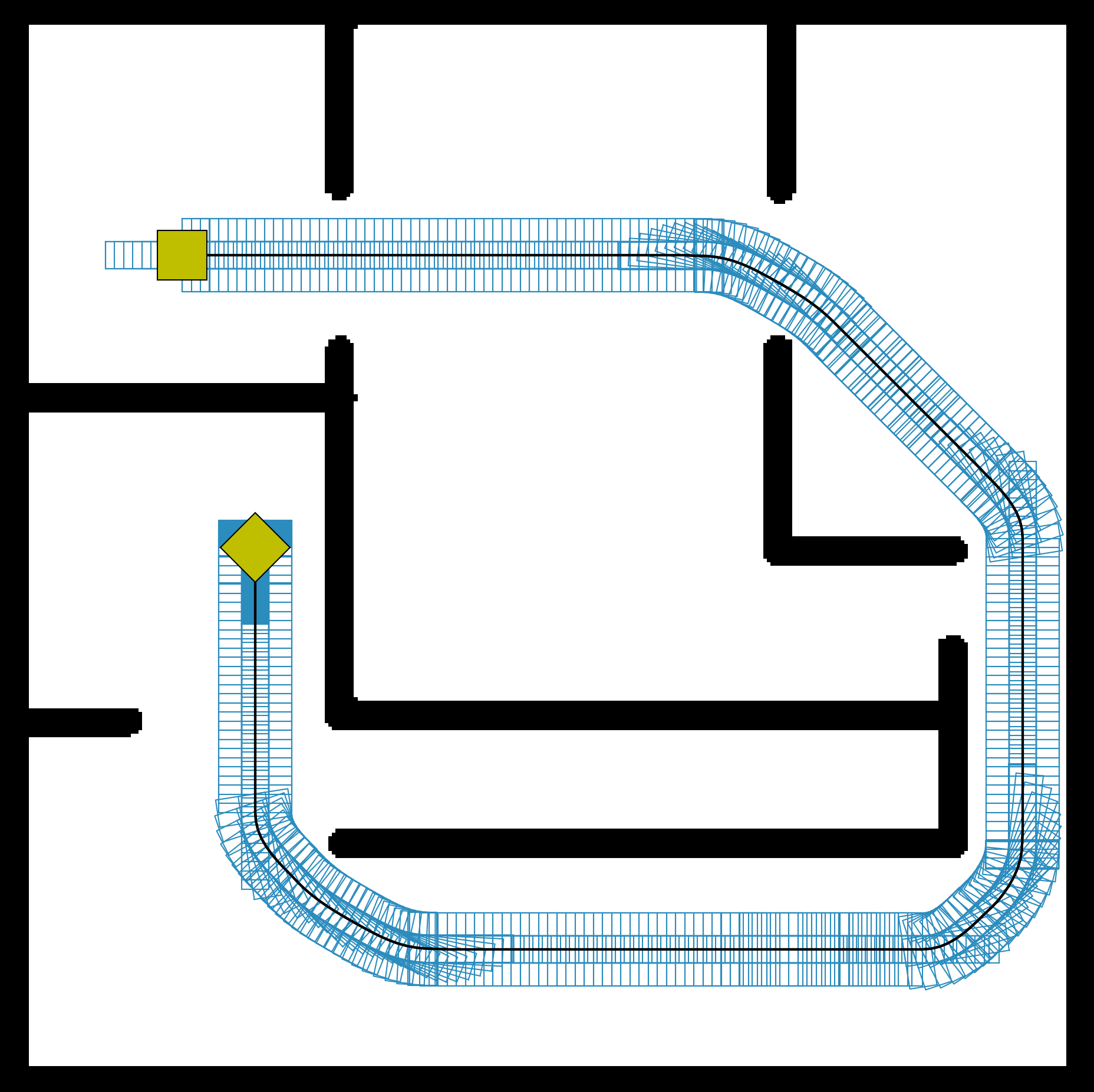} \label{fig:execution-E4}}%
    \subfigure[]{\includegraphics[height=4.8cm]{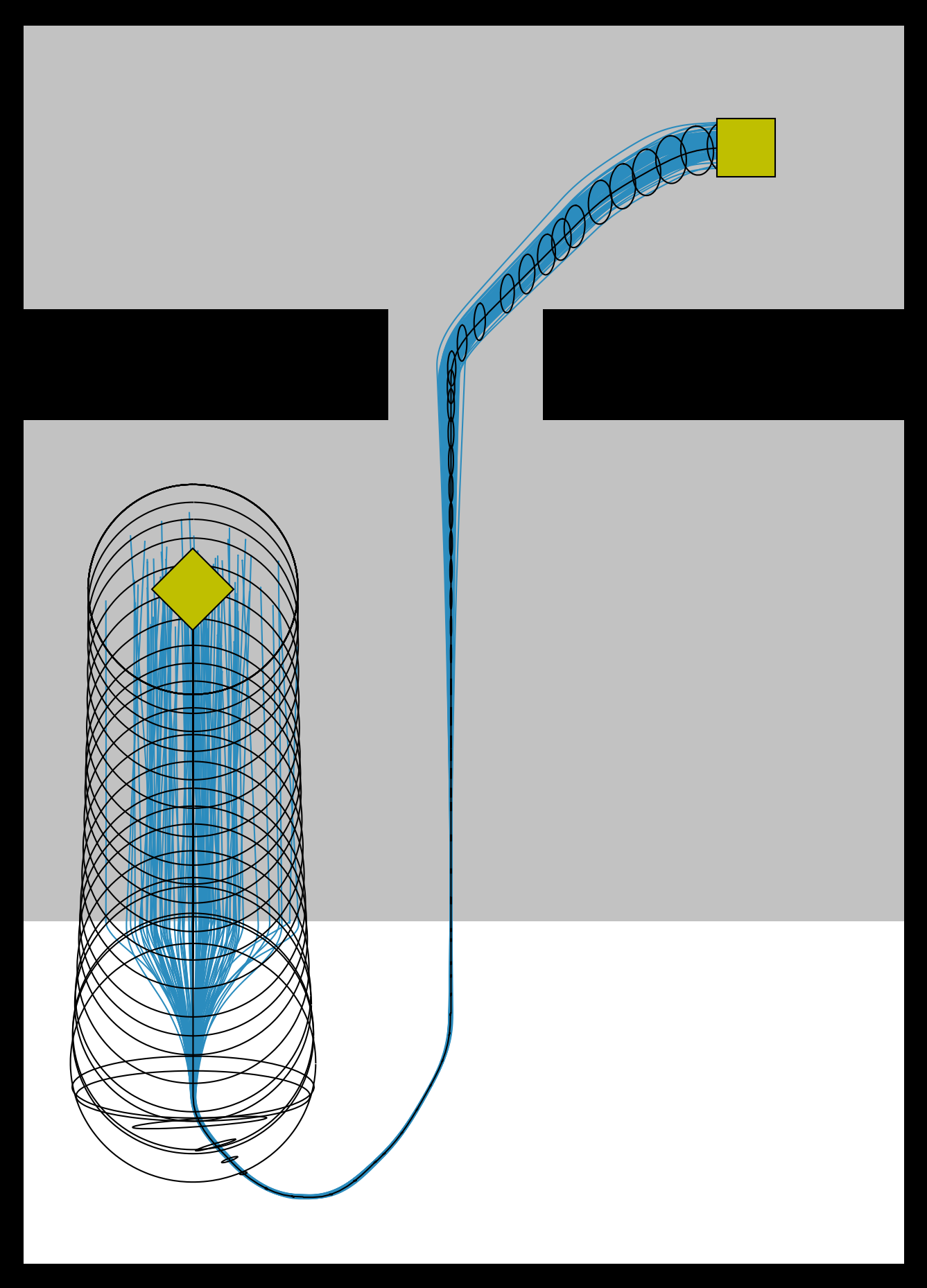} \label{fig:execution-E5}}
    
    \caption{Motion planning results. The start is the diamond, and the goal is the rectangle. (a) shows the predicted uncertainty, where the ellipses are $3 \cdot \sigma$ of the PDFs, and the location denied areas are in gray; (b) shows the path for a robot with an irregular shape (in blue).}
    
    \label{fig:executions-simulation}
\end{figure}

Fig. \ref{fig:execution-E4} shows the planned path in a $30 \times 30\ m$ squared environment. The robot has an irregular shape, like a ``T'', with dimensions $3.0 \times 0.75\ m$ and $2.0 \times 0.75\ m$ ---long and short edges. Our approach finds a safe path for this environment, whilst other approaches disregarding the real shape and the uncertainty in heading would fail. Approximating the robot by its inner circle would be unreliable due to underestimating probability of collision in the turnings. On the contrary, using a bounding box would be too conservative and some maneuvers would be discarded for being too risky ---e.g. the turning before entering the corridor.

Fig. \ref{fig:execution-E5} shows the optimal solution and the predicted PDFs for an environment of $25 \times 35\ m$ and a rectangular robot of $3.0 \times 0.75 \ m$. Due to the uncertainty, the best path in terms of probability of collision and traversal time goes outside the location denied area to receive measurements, allowing the robot to localize itself before crossing the door.

We have measured the accuracy of the estimations given by our method and compared it with that of the gamma function ---used by other approaches, like \cite{VanDenBerg2011}. To do this, we compared the output of both methods for the robot with rectangular shape, a diagonally placed obstacle, and an uncertainty of $\Sigma = I$. The error of the gamma function was inversely proportional to the distance between the robot and the obstacle, up to a 59\% for distances between $0.75\ m$ and $1.0\ m$. Conversely, as our PDF sampling strategy takes into account the deviations both in position and orientations, it obtained reliable estimations for the entire range of distances ---with only a 1.5\% of average error. To show the reliability of the planner, we run $1,000$ simulations of the solutions of Fig. \ref{fig:executions-simulation} with random noise, which were all collision free.

\begin{table}[b]
\caption{Impact of the graduated fidelity lattice.}
\label{table:graduated-fidelity}
\footnotesize\centering
\begin{tabular}{@{\extracolsep{4pt}}lrrrrr}
\cline{3-5} \cline{6-6}
& &\multicolumn{3}{c}{\textbf{Planning}} & \textbf{Solution} \\ \cline{1-2} \cline{3-5} \cline{6-6}
\textbf{Fig.} & \textbf{GF} & \textbf{Iterations} & \textbf{Insertions} & \textbf{Time (s)} & \textbf{Cost (s)} \\ \cline{1-2} \cline{3-5} \cline{6-6}
    \ref{fig:execution-E4}  & \cmark & 525  & 1,191 & 1.65 & 133.46 \\
	\ref{fig:execution-E4}  & \xmark & 2,755 & 10,844 & 11.1 & 130.86 \\
	\cline{1-2} \cline{3-5} \cline{6-6}
	\ref{fig:execution-E5} & \cmark & 1,302  & 2,868 & 2.68 & 108.27 \\
	\ref{fig:execution-E5} & \xmark & 15,952 & 42,594 & 45.38 & 95.96 \\ \cline{1-2} \cline{3-5} \cline{6-6}

\end{tabular}
\end{table}

\subsection{Planning efficiency}
To improve the planning efficiency, the proposed graduated fidelity method reduces the number of candidate paths decreasing the fidelity, but only in the uncluttered areas of the environment. Thus, high fidelity is used only when the estimated probability of collision is greater than 0 ---due to the proximity of obstacles or the amount of uncertainty. With this approach, the complex maneuvers in the control set are available where they are most needed. Similarly, using high fidelity near the goal facilitates approaching to this point.

Table \ref{table:graduated-fidelity} details the impact of the graduated fidelity lattice approach ---GF--- in the planning efficiency and performance. Columns ``Iterations'' and ``Insertions'' count the extractions and insertions in OPEN ---see Alg. \ref{alg:searchLoop}---, ``Time'' the runtime of the planner, while ``Cost'' is the traversal time of the solution. A 87.4\% average reduction in the number of iterations is achieved, although more important is the reduction in the number of insertions, which is decreased a 89.7\%. This caused the runtime to improve on average a 89.9\%, as each insertion in OPEN involves obtaining the PDFs and the probability of collision for the new candidate path. With regard to the performance, there is an average increase of a 9.7\% in the cost of the paths caused by selecting longer maneuvers in uncluttered areas, although this is almost unnoticeable when the solutions are visually compared. 

Executions in Table \ref{table:graduated-fidelity} were made without anytime search ---$\epsilon_0 = 1$--- to show the planning efficiency due to the graduated fidelity, although they improved when iteratively refining the solution until finding the optimal one. For $\epsilon_0 = 1.5$ and the experiment of Fig. \ref{fig:execution-E4}, the optimal plan was obtained in $1.2\ s$ and 422 iterations, for $\epsilon = 1.15$. For Fig. \ref{fig:execution-E5} it was obtained in $2.2\ s$ and 684 iterations, for $\epsilon = 1.03$.

These results outperform other approaches in the state of the art. In the environment of Fig. \ref{fig:execution-E5}, \cite{Bry2011} finds a valid solution in approximately $40\ s$ --- $17,500$ iterations---, while \cite{VanDenBerg2011} fails due to not expanding the paths going through the measurement area. \cite{VanDenBerg2012_IJRR} obtains in $3.65\ s$ a locally-optimal control policy based on an initial path, and their simulations show a $93\%$ of probability of success. However, this approach requires obtaining a policy for each homotopy class of trajectory to guarantee that the global optimum is found, which would lead to higher planning times.

\begin{table}[!t]
\footnotesize\centering
\caption{Efficiency of H2DMR compared to that of H2D.}
\label{table:h2dmr-performance}
\begin{tabular}{@{\extracolsep{3pt}}lrrrrrr}
\cline{2-4} \cline{5-7}       & \multicolumn{3}{c}{\textbf{Iterations}}                  & \multicolumn{3}{c}{\textbf{Time (ms)}}        \\
\cline{1-1} \cline{2-4} \cline{5-7} \multicolumn{1}{l}{\textbf{$C_{+}$}} & \multicolumn{1}{l}{\textbf{SR}} & \multicolumn{1}{l}{\textbf{MR}} & \multicolumn{1}{l}{\textbf{Gain}} & \multicolumn{1}{l}{\textbf{SR}} & \multicolumn{1}{l}{\textbf{MR}} & \multicolumn{1}{l}{\textbf{Gain}} \\ \cline{1-1} \cline{2-4} \cline{5-7}
\multicolumn{1}{l}{\textbf{$\leq$0.8}}            & 8,281                             & 3,250                               & 60.8\%                                  & 176.8                              & 81.1                       & 52.2\% \\
\multicolumn{1}{l}{\textbf{1.6}}            & 8,281                             & 842                                 & 89.8\%                                  & 123.4                              & 33.0                       & 73.3\% \\
\multicolumn{1}{l}{\textbf{3.2}}            & 8,281                             & 842                                 & 89.3\%                                  & 119.0                              & 30.0                       & 74.8\% \\
\multicolumn{1}{l}{\textbf{6.4}}            & 8,281                             & 410                                 & 95.1\%                                  & 117.8                              & 20.0                       & 83.0\% \\
\multicolumn{1}{l}{\textbf{12.8}}           & 8,281                             & 362                                 & 95.6\%                                  & 116.2                              & 19.0                       & 83.7\% \\ \cline{1-1} \cline{2-4} \cline{5-7}
\end{tabular}
\end{table}

We also compared the efficiency of H2DMR and H2D in an empty environment of $50 \times 50\ m$. To analyze the impact of the map structure in the results we used several octrees with a maximum resolution of $0.1\ m$, varying the minimum one up to $25.6\ m$. Table \ref{table:h2dmr-performance} details these results ---columns ``MR'' and ``SR'' are for H2DMR and H2D, respectively. Since H2DMR adapts the resolution to that of the octree ---with an upper bound of $f^+$---, the iterations and runtime decrease as the minimum resolution of the octree ---$C_{+}$--- increases. The optimistic nature H2DMR is not affected due to the multi-resolution grid. In fact, it has an increased connectivity ---all points between adjacent cells are linked--- and the estimations are on average a 4\% lower than those of H2D.

\subsection{Real experiments}
\begin{figure}[!b]
    \centering
    \includegraphics[width=0.9\columnwidth]{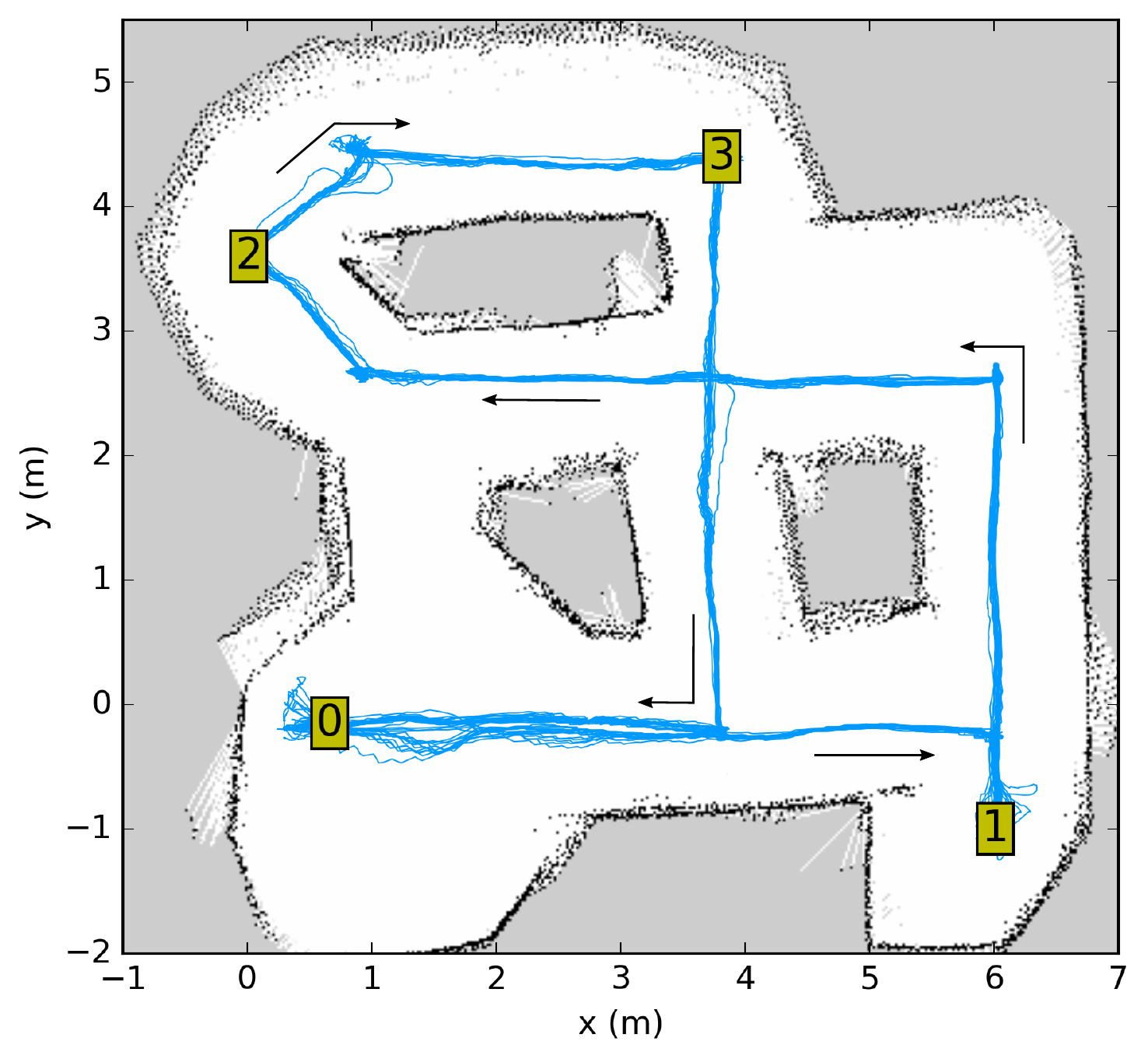}
    
    \caption{Executions in a real environment with the robot Pepper ---the robot starts in waypoint 0, then it goes to 1, 2, 3, and back to 0. Arrows indicate the direction of motion.}
    \label{fig:execution-real}
\end{figure}

In Fig. \ref{fig:execution-real} we show an experiment done with the robot Pepper (Softbank Robotics). First, we obtained the dynamics model from $512 s$ of navigation data, and then a set of 236 actions with lengths between $0.2$ and $1\ m$. The maneuvering of the robot was that of an odometric motion model, with a maximum linear speed of $0.5\ m/s$.

We used a SLAM algorithm \cite{Grisettiyz2005} to build the map, and a Monte Carlo method \cite{Fox1999} to localize the robot. The accuracy of the localization decreases when the distance to the obstacles exceeds the range of the laser sensor ---$1.5\ m$ for Pepper---, e.g. near the start or after waypoint 2. This causes deviations that are corrected by the LQG controller. Despite this, the planned path was safe, and no collisions were reported in 20  executions. Finally, the efficiency of the planner allowed the robot to operate without noticeable delays. On average, the plans between consecutive waypoints were obtained in less than 5 seconds and 1,000 iterations due to our graduated fidelity approach and the anytime search capabilities of AD*. Also, H2DMR was obtained in $0.2\ s$ ---$941$ iterations. The total cost of the path was $115\ s$.

\begin{figure}[!t]
    \centering
    \includegraphics[width=0.87\columnwidth]{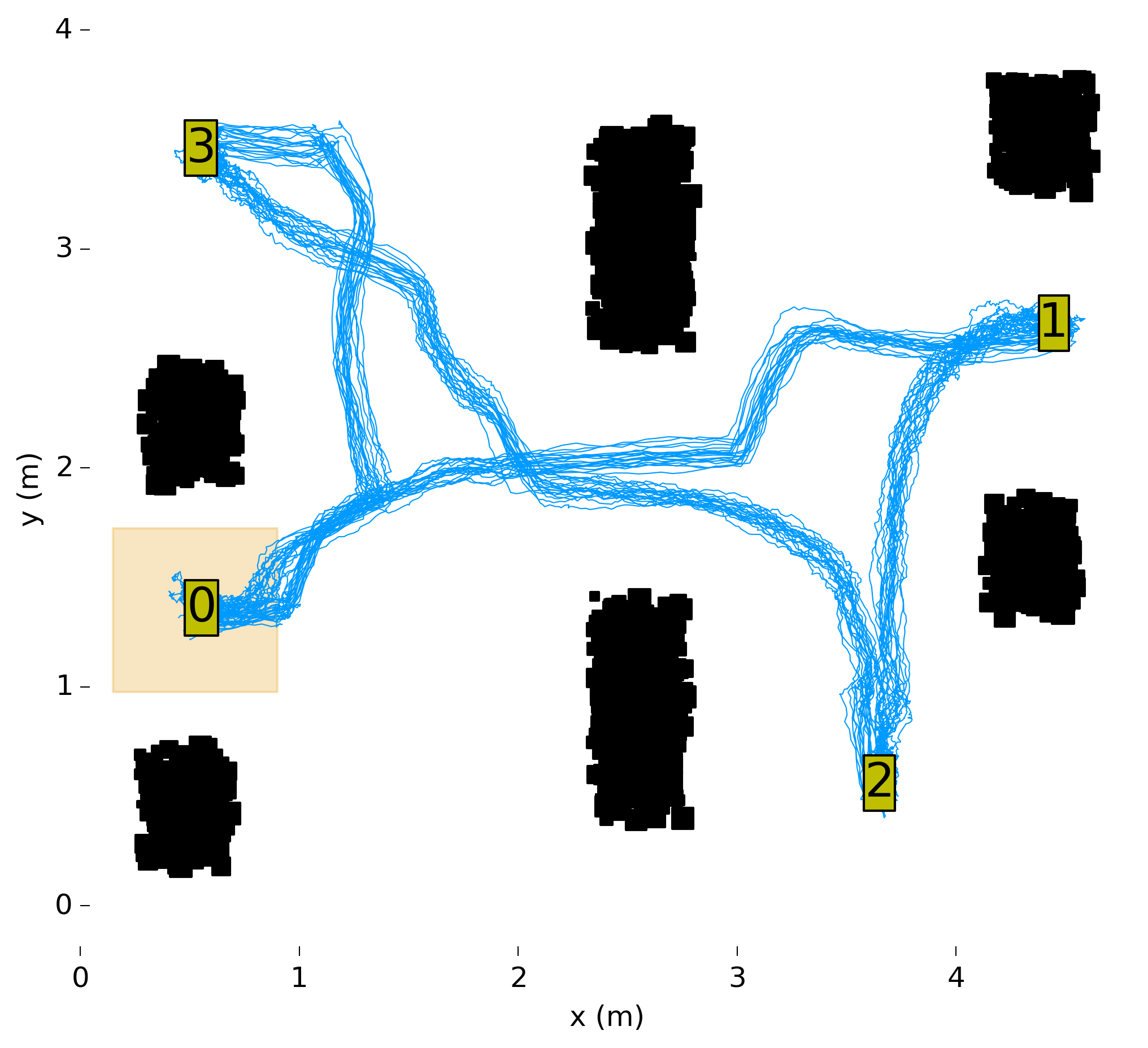}
    
    \caption{Executions in a real, cluttered environment with a UAV ---starting in waypoint 0, then going to 1, 2, 3, and back to 0. The UAV shape is in orange.}
    \label{fig:real-experiment-uav}
\end{figure}

The experiment of Fig. \ref{fig:real-experiment-uav} was done with a UAV (Asctec Pelican) localized with a motion capture system. The dynamics model was obtained from $1,458\ s$ of flight data, which resulted in a set of $3,316$ motion primitives with lengths between $0.2$ and $0.8\ m$. These motions allowed the UAV to move forward and laterally with a maximum linear speed of $0.7\ m/s$. The maximum angular speed was $30^\circ/s$. The environment was cluttered with obstacles, and the only way to navigate between the waypoints was to cross the narrow spaces between them. In fact, in some points there was only $11\ cm$ of distance between the UAV and the obstacles. Despite this, the planned path was safe, and in 20 executions no collisions were reported. H2DMR was obtained in $0.1\ s$ ---216 iterations, and on average the plans between waypoints were obtained in less than 8 seconds and 2,500 iterations. The total cost of the path was $86.9\ s$.

\section{Conclusions}
\label{sec:conclusions}
We have presented a state lattice based motion planner which manages motion and sensing uncertainty. The planner reliably obtains the probability of collision considering the real robot shape and the uncertainty in heading, and it is based on a novel graduated fidelity approach, which adapts to the obstacles in the map, and on a muli-resolution heuristic. We have reported results for simulated and real environments, with different motion models and robot shapes, proving the reliability and efficiency of the proposed method. Future research will extend the validation of the proposed algorithms in practical applications demanding autonomous navigation for UAVs. 

\bibliographystyle{IEEEtran}
\bibliography{bibliography}

\end{document}